\documentclass{article} % For LaTeX2e
\usepackage[preprint]{colm2026_conference}
\usepackage{multirow}
\usepackage{wrapfig}
\usepackage{microtype}
\usepackage{hyperref}
\usepackage{url}
\usepackage{booktabs}
\usepackage{graphicx}
\usepackage{subcaption}
\usepackage{amsmath}
\usepackage{eso-pic}
\usepackage{float}
\usepackage[table]{xcolor}

\AddToShipoutPictureBG*{ 
  \AtPageUpperLeft{
    \raisebox{-2.3cm}{  % 调整垂直位置
    \hspace{3.5cm}    % 调整水平位置
    \includegraphics[height=1.7cm]{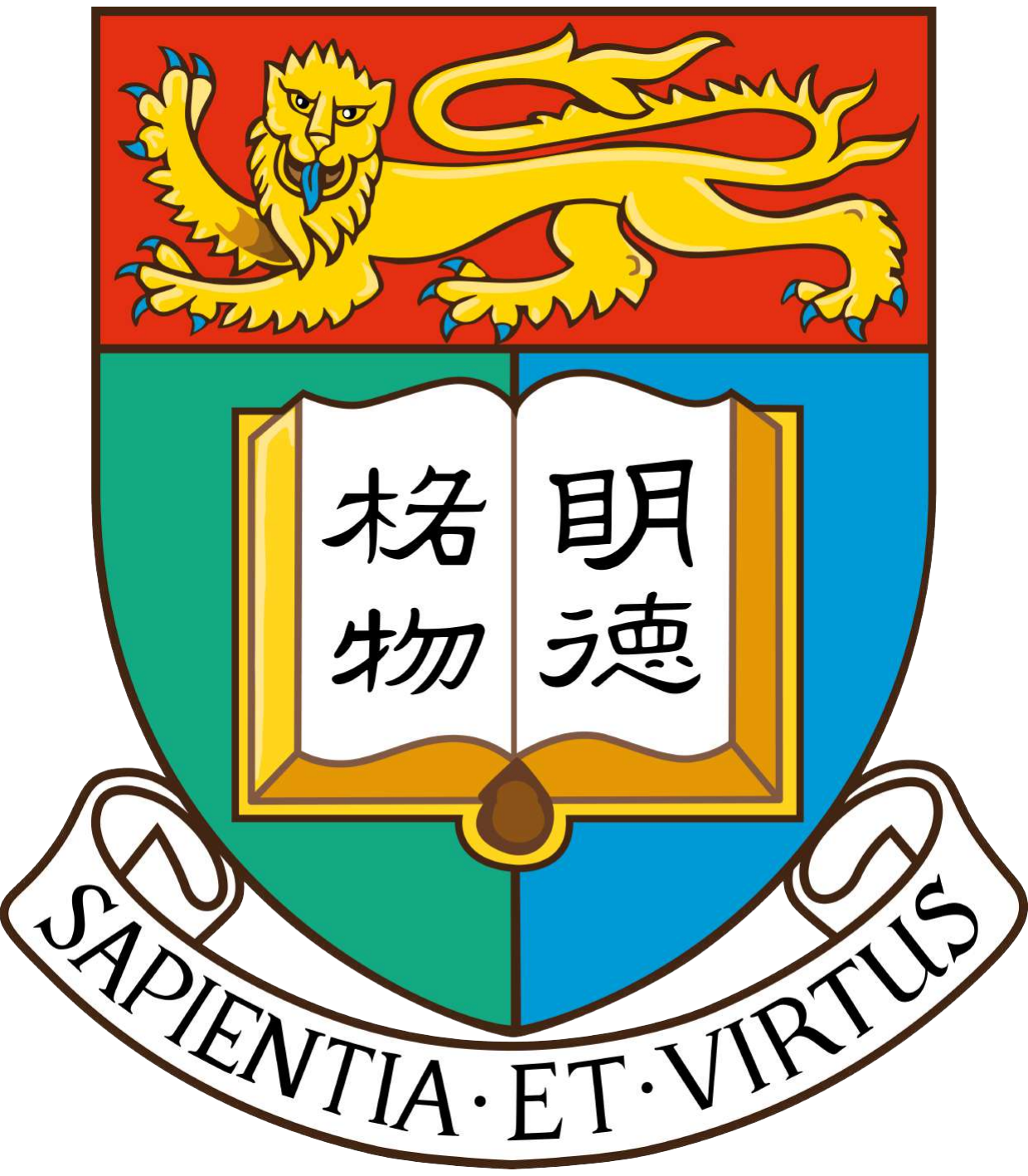}
    \hspace{10.7cm}
    \raisebox{0.2cm}{April 2026}
    }
  }
}

\usepackage{lineno}

\definecolor{darkblue}{rgb}{0, 0, 0.5}
\hypersetup{colorlinks=true, citecolor=darkblue, linkcolor=darkblue, urlcolor=darkblue}

\title{CodeComp: Structural KV Cache Compression for Agentic Coding}

\author{
  \textbf{Qiujiang Chen}$^{1}$\thanks{Contact Email: qiujiangc@outlook.com, Equal contribution.},\;
  \textbf{Jing Xiong}$^{1}$\footnotemark[1],\;
  \textbf{Chenyang Zhao}$^{2}$\; 
  \textbf{Sidi Yang}$^{1}$,\;
  \textbf{Ngai Wong}$^{1}$ 
  \\[0.6em]
  \normalfont\small $^{1}$The University of Hong Kong\qquad
  $^{2}$LMSYS Org
}

\begin{document}
\ifcolmsubmission
\linenumbers
\fi

\vspace*{-0.5cm}
% \vspace{-1cm}
{\color{gray}\noindent\rule{\textwidth}{0.4pt}}
\vspace*{-0.1cm}
% \vspace{0.02cm}
\maketitle
\thispagestyle{empty}
% \vspace{-0.5cm}
\begin{abstract}
Agentic code tasks such as fault localization and patch generation require processing long codebases under tight memory constraints, where the Key-Value (KV) cache becomes the primary inference bottleneck. Existing compression methods rely exclusively on attention signals to estimate token importance, systematically discarding structurally critical tokens such as call sites, branch conditions, and assignments that are essential for code understanding. We present CodeComp, a training-free KV cache compression framework that incorporates static program analysis into LLM inference via Code Property Graph priors extracted by Joern. Across bug localization and code generation benchmarks, CodeComp consistently outperforms attention-only compression baselines under equal memory budgets, recovering the majority of full-context accuracy 
under aggressive KV cache compression, while matching the patch generation quality of uncompressed full-context inference and integrating seamlessly into SGLang-based agentic coding pipelines without model modification. 
% Code is available at \url{https://anonymous.4open.science/r/CodeComp}.
\end{abstract}

\section{Introduction}
Agentic code tasks, such as fault localization and patch generation across entire codebases, are among the most context-demanding applications of Large Language Models (LLMs) in software engineering.
LLMs have shown strong capabilities on these tasks, but efficient deployment at repository scale is constrained by a primary bottleneck: the Key-Value (KV) cache.
This cache stores intermediate attention states to avoid redundant computation, but it grows linearly with context length and can consume up to 70\% of total GPU memory during long-context inference~\citep{liu2025chunkkv}.
For repository-level inputs spanning tens of thousands of tokens, this memory overhead quickly exceeds the capacity of available hardware~\citep{devoto2025expectedattentionkvcache}.
\begin{figure}[H]
    \centering
    \includegraphics[width=0.9\linewidth, height=4cm]{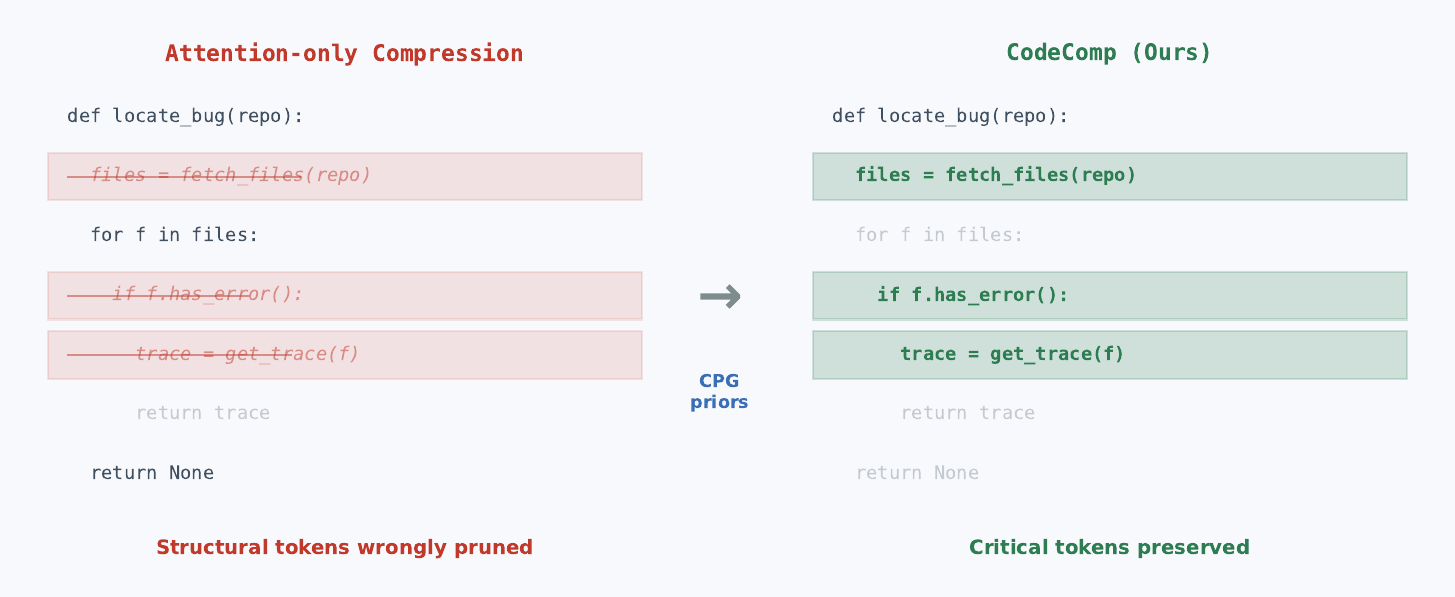}
    \caption{Attention-only compression discards structurally 
    critical tokens such as function calls, branch predicates, 
    and return statements. CodeComp uses CPG priors to 
    explicitly preserve these tokens.}
    \label{fig:teaser}
\end{figure}
A growing body of work addresses this challenge through KV cache compression. Token eviction methods, such as StreamingLLM~\citep{xiao2023efficient}, H2O~\citep{zhang2023h2o}, and SnapKV~\citep{li2024snapkv}, retain tokens with high attention scores. Quantization approaches, including KIVI~\citep{liu2024kivi} and KVQuant~\citep{hooper2024kvquant}, reduce the bit-width of cache entries. More recent methods, such as ChunkKV~\citep{liu2025chunkkv} and ParallelComp~\citep{xiong2025parallelcompparallellongcontextcompressor}, operate on chunks instead of individual tokens, enabling more efficient parallel processing of long contexts.
Despite their diversity, these approaches largely share a common design principle: \textit{token importance is inferred primarily from attention signals}.

This is a reasonable assumption for natural language, but it breaks down for source code. Source code differs fundamentally from natural language in that its semantics are governed by formal program structure (e.g., control flow, data dependencies, and inter-procedural call relationships) rather than surface-level token co-occurrence. 
We adopt a standard retrieval pipeline for context selection with perplexity-based re-ranking, and focus our technical contribution on the compression stage, where two fundamental challenges arise.

% \textbf{C1: Lexical retrieval has limited recall for repository-level code tasks.} Existing pipelines rely on lexical retrieval methods (e.g., BM25) to select relevant files from a repository~\citep{xia2024agentlessdemystifyingllmbasedsoftware}. However, code-relevant files are often lexically distant from the query, yet structurally connected through call chains or shared data dependencies that text-similarity signals fail to capture — a limitation well documented in prior work~\citep{jimenez2024swebenchlanguagemodelsresolve}. In this work, we adopt a two-stage selection pipeline (initial retrieval followed by perplexity-based re-ranking) as a practical approximation, and focus our technical contribution on the compression stage.

\textbf{C1: Attention-centric compression discards structurally critical tokens.} Even when relevant chunks are identified, existing KV cache compression methods, such as token eviction~\citep{zhang2023h2o, li2024snapkv} and chunk-level parallel compression~\citep{xiong2025parallelcompparallellongcontextcompressor}, estimate token importance purely from attention scores. This approach fails to capture deeper semantic significance. For example, variable definitions at the head of a def-use chain, branch conditions governing faulty logic, and call sites linking cross-file dependencies often receive low attention at a given decoding step, yet are essential for correct localization.
Attention is a learned, query-driven proxy that does not necessarily align with program-semantic importance.

\textbf{C2: Optimal budget allocation should be structure-aware.} 
In agentic code tasks, different chunks vary greatly in their structural relevance to the query. Allocating a uniform KV cache budget across chunks wastes capacity on less relevant regions while under-representing structurally critical ones. Existing methods rely solely on attention statistics or uniform allocation, neither of which reflects program-structural importance when distributing 
the compression budget.

% \textbf{C2: Compression budget allocation should be structure-aware and query-conditioned.} In repository-level code tasks, different chunks vary greatly in their structural relevance to the query. Allocating a uniform KV cache budget across chunks wastes capacity on less relevant files while under-representing structurally critical ones. To our knowledge, no existing method allocates compression budget according to structural importance. We incorporate this principle into CodeComp, though we note that its marginal contribution relative to span-level protection is dataset-dependent, as analyzed in \S5.

Together, these challenges suggest that effective KV cache compression for agentic code tasks requires incorporating program structure beyond attention signals.

We propose \textbf{CodeComp}, a training-free KV cache compression framework for agentic code tasks that incorporates structural priors from static program analysis. CodeComp leverages the Code Property Graph (CPG)~\citep{CPG} extracted by Joern~\citep{joern_github} to guide two complementary compression decisions: \textbf{span-level structural protection}, which preserves structurally critical tokens from eviction, and \textbf{structure-aware budget allocation}, which distributes the compression budget according to each chunk's structural importance. Implemented on SGLang~\citep{zheng2024sglangefficientexecutionstructured}, CodeComp integrates seamlessly into large-scale agentic code workflows without modifying the underlying model.

Our contributions are as follows:
\begin{itemize}
    \item We observe that existing KV cache compression methods exhibit a structural blind spot when applied to code: attention-based importance estimation fails to capture program-semantic relationships, leading to systematic mis-pruning of structurally critical tokens (e.g., callsites, branches, and def-use relations) (\S\ref{sec:motivation}).
    \item We propose \textbf{CodeComp}, the first KV cache compression framework to incorporate static program analysis into KV cache compression, using CPG priors from Joern to guide both span-level protection of semantically critical tokens and structure-aware budget allocation across chunks (\S\ref{sec:method}).
    \item Empirically, CodeComp consistently outperforms attention-only compression baselines across bug localization and code generation benchmarks under the same memory budget, recovering the majority of full-context accuracy under aggressive KV cache compression and matching the patch generation quality of uncompressed full-context inference (\S\ref{sec:experiment}).
\end{itemize}
\section{Related Work}
\textbf{KV Cache Compression for Efficient Inference.}
The rapid growth of context length in large language models has made KV cache memory a major bottleneck during inference. A substantial body of work has been proposed to reduce KV cache footprint through compression. Token-level eviction methods such as StreamingLLM~\citep{xiao2023efficient}, H2O~\citep{zhang2023h2o}, and SnapKV~\citep{li2024snapkv} selectively retain tokens based on attention scores or related heuristics. Complementary to eviction, quantization-based approaches including KIVI~\citep{liu2024kivi} and KVQuant~\citep{hooper2024kvquant} reduce the precision of stored key-value pairs to lower memory usage. More recent work explores chunk-level or parallel compression strategies, such as ChunkKV~\citep{liu2025chunkkv} and ParallelComp~\citep{xiong2025parallelcompparallellongcontextcompressor}, which operate on larger units to improve scalability for long contexts.
Despite differences in granularity and implementation, most existing methods estimate token importance primarily based on attention signals. While such signals are effective in general language modeling settings, they do not explicitly account for the structural dependencies inherent in source code, where semantic relevance is often determined by program structure rather than token co-occurrence.

\textbf{LLM for Agentic Code Tasks.}
Large language models have demonstrated strong performance on code-related tasks, including code generation, program understanding, and bug fixing~\citep{rozière2024codellamaopenfoundation, li2023starcodersourceyou, nijkamp2023codegenopenlargelanguage}. Recent benchmarks such as SWE-bench~\citep{jimenez2024swebenchlanguagemodelsresolve} and LCA~\citep{bogomolov2024long} highlight the challenges of repository-level reasoning, where models must process large codebases and reason over long, structurally complex contexts. Despite progress, performance on these tasks remains far below human level, with even the strongest models struggling when relevant context is sparse, cross-file dependencies are deep, or the KV cache budget is constrained. Existing pipelines typically address context scale through retrieval~\citep{xia2024agentlessdemystifyingllmbasedsoftware}, but do not consider how to compress the retrieved context in a structure-aware manner. However, none of these methods consider how to compress the retrieved context in a structure-aware manner, leaving a gap that we address in this work.

\textbf{Program Analysis and Structural Representations.}
Prior work in program analysis has introduced structural representations such as ASTs, CFGs, and PDGs to capture syntax, control flow, and data dependencies beyond the lexical form of code. CPG unifies these representations into a single graph, enabling joint reasoning across multiple structural dimensions. Tools such as Joern~\citep{joern_github} provide practical implementations for extracting CPGs from real-world codebases.
Structural representations have been widely used in tasks such as vulnerability detection, program understanding, and graph-based code modeling~\citep{chen2025bridgingcodegraphslarge}. However, their use has largely been limited to model training or static analysis pipelines. To our knowledge, this is the first work to incorporate static program analysis into KV cache compression, enabling structure-aware preservation of semantically critical tokens in code.
\section{Motivation}
\label{sec:motivation}

Existing KV cache compression methods primarily rely on attention scores to estimate token importance. In natural language, this assumption breaks down in code, where semantic correctness depends on structured program elements rather than token-level saliency. We therefore ask: \emph{what exactly is lost when compression relies only on attention?}

\begin{figure*}[t]
    \centering

    \begin{subfigure}[t]{0.32\textwidth}
        \centering
        \includegraphics[width=\linewidth]{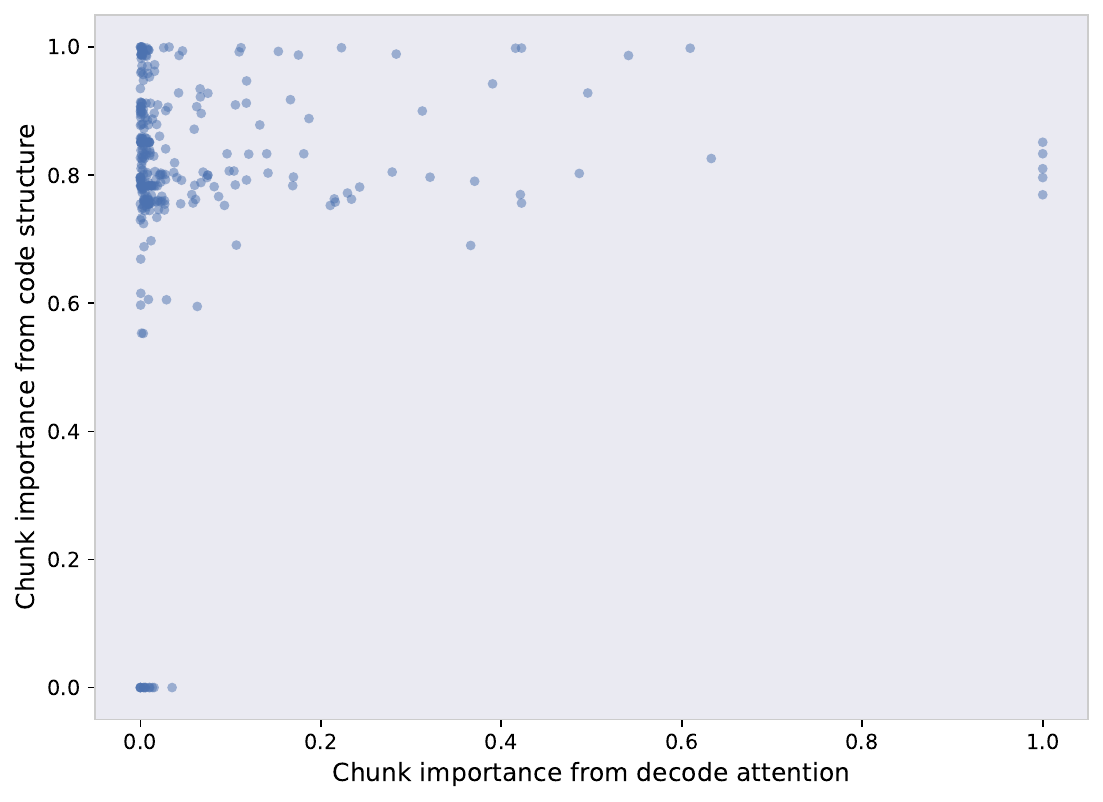}
        \caption{Attention vs. structure}
    \end{subfigure}
    \hfill
    \begin{subfigure}[t]{0.32\textwidth}
        \centering
        \includegraphics[width=\linewidth]{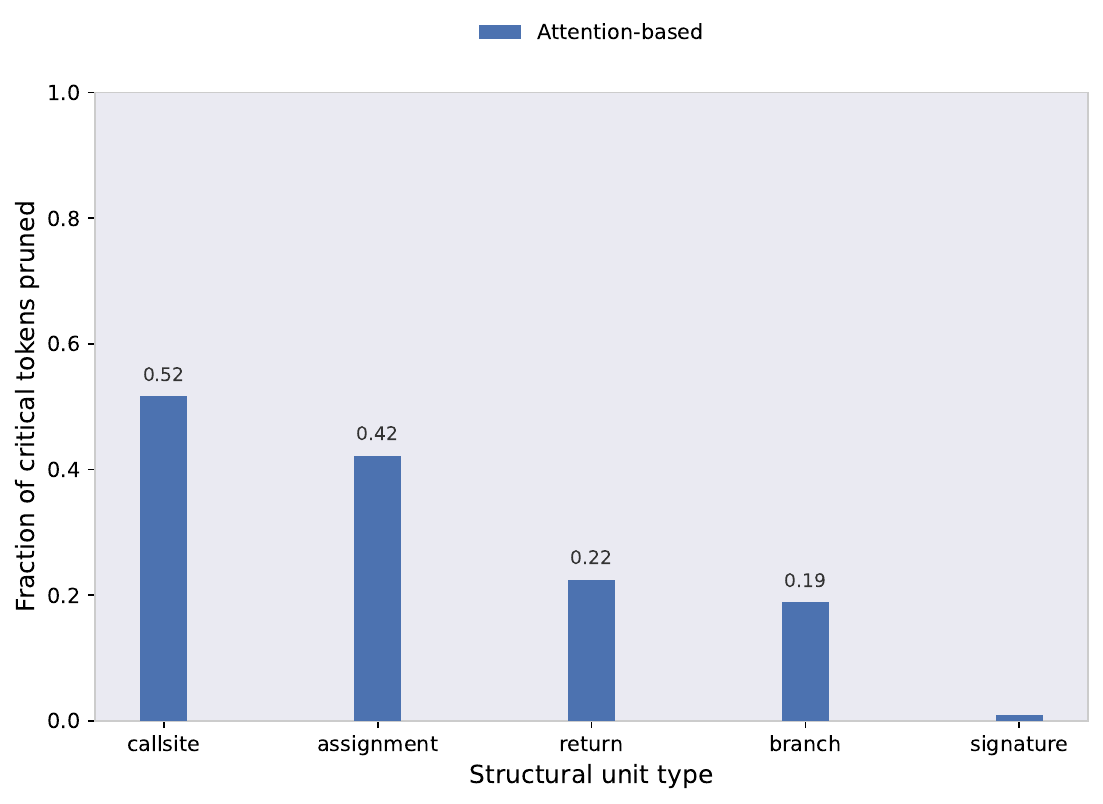}
        \caption{Structural mis-pruning}
    \end{subfigure}
    \hfill
    \begin{subfigure}[t]{0.32\textwidth}
        \centering
        \includegraphics[width=\linewidth]{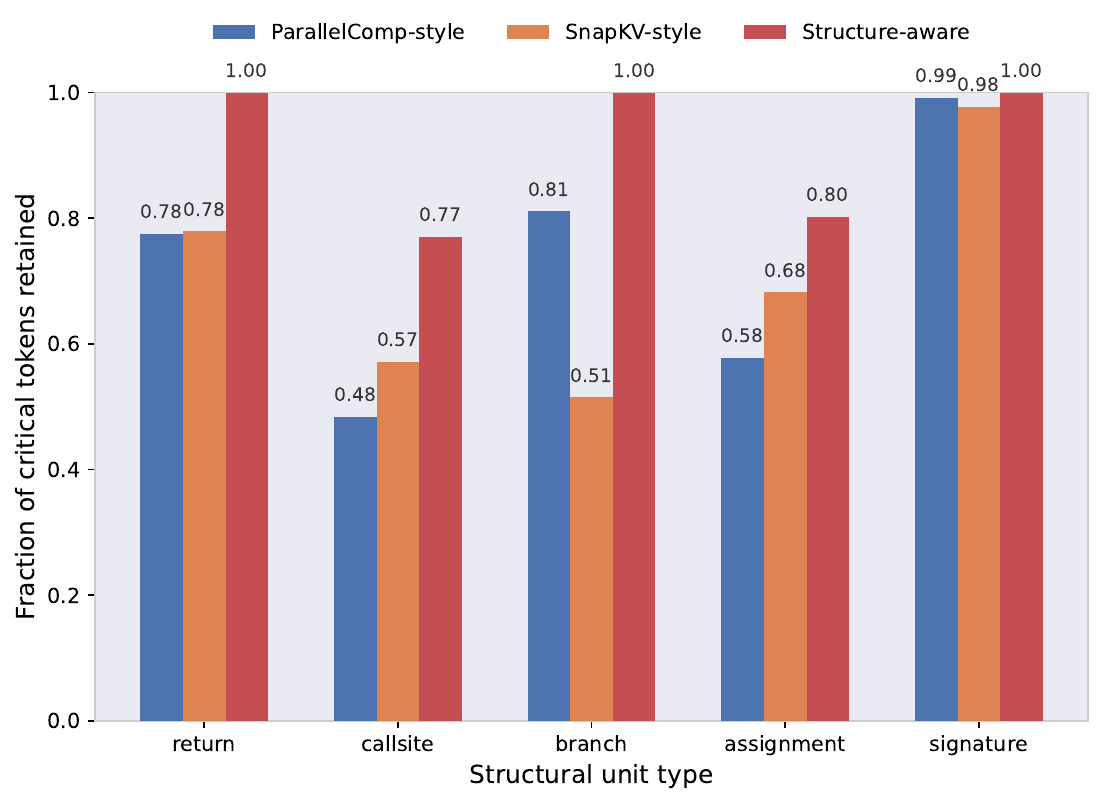}
        \caption{Structure-aware retention}
    \end{subfigure}

    \caption{
    Motivation analysis for KV cache compression in code.
    (a) Attention--structure mismatch: attention-based importance is weakly aligned with CPG-derived structural importance, causing structurally critical chunks to receive low attention scores.
    (b) Structural mis-pruning: under attention-only compression, a large fraction of critical structural tokens (e.g., callsite, assignment, and return) are incorrectly discarded.
    (c) Structure-aware retention: incorporating structural priors significantly improves the preservation of critical structural tokens across categories compared to attention-based methods.
    }
    \label{fig:motivation}
\end{figure*}
\textbf{Attention--structure mismatch.}
For each chunk, we compute an attention-based importance score and compare it with a CPG-derived structural score. As shown in Figure~\ref{fig:motivation}(a), the two signals are largely misaligned: the Jaccard overlap between the top-20\% chunks ranked by attention and those ranked by structural importance is only $0.0944$, indicating that the two signals select largely disjoint sets of chunks. This suggests that attention scores are not a reliable proxy for program-semantic importance in code.

\textbf{Structural mis-pruning.}
This mismatch leads to systematic errors during compression.
Figure~\ref{fig:motivation}(b) shows the fraction of structurally critical tokens pruned under attention-only compression across different structural categories, including callsite, assignment, return, branch, and signature. 
A substantial portion of these tokens are incorrectly discarded, demonstrating that attention-based selection tends to remove semantically critical program elements rather than unimportant tokens.

\textbf{Structure-aware retention.}
We next examine whether incorporating structural signals can mitigate this issue.
Figure~\ref{fig:motivation}(c) compares the retention of critical structural tokens under different selection strategies, including the attention-based approach (SnapKV, ParallelComp) and the structure-aware approach (CodeComp). The structure-aware approach consistently preserves a higher fraction of critical structural tokens across all categories. For example, on callsite tokens, CodeComp retains 1.00 compared to 0.48 for ParallelComp and 0.52 for SnapKV, demonstrating that structural priors effectively reduce mis-pruning and better maintain program semantics under compression.

\begin{wraptable}{r}{0.58\columnwidth}
\vspace{-6pt}
\hspace{-5pt}
\centering
\small
\setlength{\tabcolsep}{4pt}
\begin{tabular}{lcccc}
\toprule
Method & API Prec & API Rec & API Jac & Edit Dist \\
\midrule
ParallelComp & 0.1318 & 0.1536 & 0.0861 & 0.7046 \\
SnapKV & 0.0196 & 0.0909 & 0.0175 & 0.8615 \\
CodeComp & \textbf{0.1929} & \textbf{0.2975} & \textbf{0.1471} & \textbf{0.5998} \\
\bottomrule
\end{tabular}
\vspace{4pt}
\caption{KV cache compression results on LCA.}
\label{tab:motivation_results}
\vspace{-10pt}
\end{wraptable}
We further evaluate how these differences affect end-to-end code generation.
Under identical chunk selection and compression budgets, we compare attention-based methods with the structure-aware approach on the LCA~\citep{bogomolov2024long} (Library-based Code Generation) benchmark. As shown in Table~\ref{tab:motivation_results}, attention-only methods achieve substantially lower API precision, recall, and Jaccard similarity, along with higher normalized edit distance. In contrast, the structure-aware method consistently improves all metrics, producing outputs that are both more API-correct and closer to the reference solution.

\textbf{Key insight.} These findings reveal that attention-based compression exhibits a structural blind spot in code: it systematically mis-prunes program-semantic evidence that is critical for downstream code tasks, motivating the structure-aware compression mechanisms in CodeComp.
\section{Method}
\label{sec:method}

\begin{figure*}[h]
    \centering
    \includegraphics[width=\textwidth]{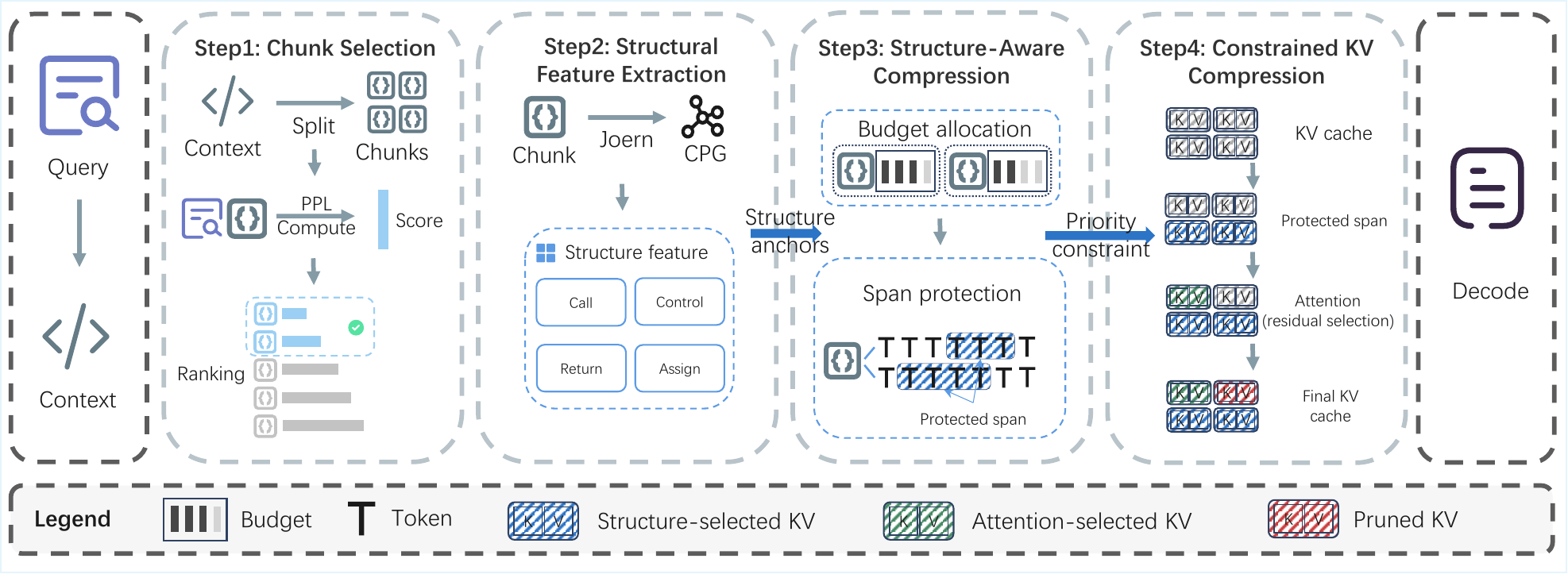}
    \caption{
    Overview of \textbf{CodeComp}, a structure-aware KV cache compression framework. Given a query and retrieved repository context, we first select relevant chunks via PPL-based scoring (Step~1). We then extract structural anchors from static program analysis using Joern and CPG, identifying semantically critical code elements including function calls, control-flow predicates, return statements, and assignments (Step~2). These anchors are used to allocate chunk-level compression budgets and protect semantically critical spans (Step~3). Finally, attention-based compression fills the remaining capacity under these constraints (Step~4).
    }
    \label{fig:overview}
\end{figure*}

\textbf{Overview.}
CodeComp incorporates structural signals from static program analysis into KV cache compression for code-centric tasks.
Given a query and repository-level context, we first perform query-conditioned chunk selection using perplexity-based filtering (\S\ref{sec:chunk_selection}).
For the selected chunks, we extract structural signals from CPG-derived features (\S\ref{sec:structural_extraction}).
These signals are used to allocate compression budgets across chunks (\S\ref{sec:budget_allocation}) and to preserve semantically critical local spans in each chunk (\S\ref{sec:span_protection}).
Finally, attention-based token selection is applied within these structural constraints to fill the remaining capacity (\S\ref{sec:attention_compression}).

\subsection{Query-Conditioned Chunk Selection}
\label{sec:chunk_selection}

Given a repository-level input, we partition the candidate code context into a set of chunks $\{C_i\}$.
Whenever possible, we use structure-aware boundaries such as method- or function-level units to preserve semantic coherence.
Long methods are further split to satisfy the token budget, while overly short fragments are merged with neighboring chunks.

We score each chunk using query-conditioned perplexity (PPL).
For each chunk $C_i$, we evaluate the likelihood of the query conditioned on $\text{prefix} + C_i$:
\begin{equation}
s_i^{\text{ppl}} = \frac{1}{T} \sum_{t=1}^{T} -\log p(q_t \mid \text{prefix}, C_i, q_{<t}).
\end{equation}
We then select the top-$k$ chunks with the lowest perplexity:
\begin{equation}
\mathcal{S} = \operatorname{TopK}_{i}(-s_i^{\text{ppl}}),
\end{equation}
where $\mathcal{S}$ denotes the set of selected chunks.
This stage reduces the search space to a small set of query-relevant chunks before structural information is introduced.

\subsection{Structural Feature Extraction}
\label{sec:structural_extraction}
 
For each selected chunk, we extract structural signals from its CPG to characterize semantically meaningful program events. Unlike lexical retrieval methods such as BM25, which measure query-document similarity based on token co-occurrence, these signals are derived from program structure and capture code-specific semantics such as control flow, data dependencies, and inter-procedural interactions, independent of surface-level token frequency.

\textbf{Intra-chunk structural features.}
For each chunk $C_i$, we collect structural statistics from its CPG subgraph, including node-level features such as function calls, control-flow constructs, return statements, and assignments, as well as edge-level features such as control-flow and program-dependence edges:
\begin{equation}
\mathbf{f}_i = \big(N_{\text{call}}, N_{\text{control}}, N_{\text{return}}, N_{\text{assign}}, E_{\text{cfg}}, E_{\text{pdg}}\big).
\end{equation}
% Function calls mark API interactions and execution entry points, control-flow predicates encode decision logic, return statements expose program outcomes, and assignments indicate state updates.
% These features define the structural anchors used in both budget allocation and span construction.

\textbf{Normalization.}
To make feature values comparable across chunks of different sizes, we apply logarithmic normalization:
\begin{equation}
\operatorname{Norm}(x; \tau) = \min\left(1,\; \frac{\log(1+x)}{\log(1+\tau)}\right),
\end{equation}
where $\tau$ is a feature-specific scaling constant.
The normalized structural score for chunk $C_i$ is:
\begin{equation}
\sigma_i = \sum_{k} w_k \cdot \operatorname{Norm}(f_{i,k}; \tau_k).
\end{equation}

\subsection{Structure-Aware Budget Allocation}
\label{sec:budget_allocation}

We use the chunk-level structural score $\sigma_i$ to modulate compression budgets across selected chunks, so that structurally more important chunks receive larger capacity.

We normalize scores across selected chunks:
\begin{equation}
\tilde{s}_i =
\begin{cases}
0.5, & \text{if } \sigma_{max} = \sigma_{min}, \\
\frac{\sigma_i - \sigma_{min}}{\max(\epsilon,\; \sigma_{max} - \sigma_{min})}, & \text{otherwise},
\end{cases}
\end{equation}
and map each normalized score to a capacity multiplier:
\begin{equation}
m_i = m_{min} + (m_{max} - m_{min}) \cdot \operatorname{clip}(\tilde{s}_i, 0, 1).
\end{equation}
The token budget for chunk $C_i$ is then:
\begin{equation}
B_i = \left\lfloor |C_i| \cdot \min(r_{max},\; r \cdot m_i) \right\rfloor,
\end{equation}
where $r$ is the base compression ratio.

\subsection{Span-Level Structural Protection}
\label{sec:span_protection}

Beyond allocating budgets at the chunk level, we introduce span-level structural protection to explicitly preserve semantically critical local code regions within each chunk.
Chunk-level allocation determines how much context is retained, but not which specific tokens survive compression.
For code-centric tasks, decisive evidence is often concentrated in a small number of local regions such as call expressions, control-flow predicates, and return statements.
Span-level protection ensures these regions are preserved regardless of their attention scores.

\textbf{Structural span construction.}
We define a \textit{structural span} as a contiguous token sequence associated with a semantically meaningful program element in the CPG, such as a function signature, call expression, control-flow predicate, return statement, or assignment.
For each selected chunk $C_i$, we extract a set of candidate spans $\mathcal{Z}_i = \{z_1, z_2, \dots\}$.
Each span is anchored by one structural event and optionally extended with a small local context window, so that each retained span forms a compact and semantically interpretable local reasoning unit.

\textbf{Span scoring.}
We assign each span a structural importance score:
\begin{equation}
s(z) = \sum_{k} w_k f_k(z),
\end{equation}
where $f_k(z)$ are binary indicators of structural types (call, control, return, assignment).
This scoring function prioritizes spans corresponding to semantically meaningful program events, operating over structurally defined local units rather than isolated tokens.

\textbf{Hard query-aware protection.}
We additionally apply a hard protection rule to spans that overlap with symbols appearing in the query.
Let $S(z)$ denote the symbols in span $z$ and $Q$ denote the query symbols.
We define:
\begin{equation}
p^{\text{query}}(z) =
\begin{cases}
1, & \text{if } |S(z) \cap Q| > 0, \\
0, & \text{otherwise.}
\end{cases}
\end{equation}
Spans with $p^{\text{query}}(z)=1$ are treated as protected spans, ensuring that query-relevant local code regions are retained even when their structural score is not high.

\textbf{Span selection under budget.}
We reserve a portion of the chunk budget for span selection:
\begin{equation}
B_i^{\text{span}} = \min\big(B_i,\; \max(B_{min},\; \lfloor \rho_{\text{span}} B_i \rfloor)\big).
\end{equation}
Spans are selected in two stages:
(1) protected spans (signatures and query-matched spans) are included first;
(2) remaining spans are ranked by $s(z)$ and added greedily until the span budget is exhausted.

\textbf{Token-level protection.}
Let $P_i$ denote the tokens covered by selected spans.
If $|P_i| \ge B_i$, we retain the first $B_i$ tokens in $P_i$.
Otherwise, we retain all tokens in $P_i$ and fill the remaining budget with tokens nearest to $P_i$ in the original sequence, preserving local syntactic continuity around structurally important regions.

\subsection{Attention-based KV Cache Compression}
\label{sec:attention_compression}

After span-level protection determines which tokens must be retained, attention-based compression fills the remaining capacity within each chunk.

During the prefill stage, let $Q \in \mathbb{R}^{W \times d}$ and $K \in \mathbb{R}^{L_c \times d}$ denote the query and context key representations within a window of size $W$.
We compute per-token importance scores:
\begin{equation}
A = \operatorname{softmax}\left(\frac{QK^\top}{\sqrt{d}}\right),
\qquad
u(j) = \sum_{t=1}^{W} A(t, j).
\end{equation}
We retain the top-ranked context tokens according to $u(j)$ up to the remaining chunk budget.
Compression is applied independently at each transformer layer.
Structural priors thus determine the critical evidence that must be preserved, while attention-based compression serves as a residual selector for the remaining capacity.

\subsection{Position Encoding}
\label{sec:position_encoding}

\textbf{PPL-based chunk scoring.}
During the chunk selection stage, each candidate chunk $C_i$ is scored independently. The input to the model is formatted as $[\text{prefix}; C_i; \text{query}]$, where position indices are assigned contiguously starting from 0. This allows parallel scoring of all chunks without positional interference between them.

\textbf{Compression and decode.}
After top-$k$ chunks are selected and compressed, the final input is assembled as:
\begin{equation}
[\text{prefix};\; \hat{C}_1;\; \hat{C}_2;\; \dots;\; 
\hat{C}_k;\; \text{query}]
\end{equation}
where $\hat{C}_i$ denotes the compressed version of chunk $C_i$. Each chunk retains its original position indices from the scoring stage. Since chunks are scored independently, their position indices may overlap. To avoid positional conflict with the query, we assign the query a starting 
position of:
\begin{equation}
p_{\text{query}} = \max_{i} \operatorname{len}(C_i) + 
\operatorname{len}(\text{prefix})
\end{equation}
where $\operatorname{len}(C_i)$ is the token length of chunk 
$C_i$ before compression. This ensures that the query always receives the highest position indices in the sequence, consistent with its role as the decoding target, regardless of the actual lengths of the compressed chunks.
\section{Experiment}
\label{sec:experiment}
We evaluate CodeComp on two representative code tasks: code generation and bug localization. Detailed experimental settings and hyperparameters are provided in the Appendix.
% \subsection{Code Generation}
% Table~\ref{tab:swebench_lite} reports results on SWE-bench Lite. Under both compression settings, CodeComp consistently outperforms ParallelComp across the two evaluated code generation models, yielding substantially better ground-truth file-level precision, recall, F1, and Jaccard similarity, while also producing more valid patches and lower edit distance in most cases. The improvements are particularly pronounced for DeepSeek-Coder, where naive compression causes severe performance degradation, whereas CodeComp preserves much more of the original generation quality. These results further support that structure-aware KV compression is effective for repository-level code generation, especially when the model must retain semantically critical project context under limited cache budgets.

\subsection{Code Generation}
Table~\ref{tab:swebench_lite} reports results on SWE-bench Lite~\citep{jimenez2024swebenchlanguagemodelsresolve}.
CodeComp consistently outperforms ParallelComp across both models and compression budgets.
The improvements are most pronounced for DS-Coder: at cap=0.4, ParallelComp achieves GF F1 of only 0.021 (\textbf{16$\times$} below the no-compression baseline), while CodeComp recovers to 0.250 with patch validity of 1.000.
On Qwen-Coder, CodeComp achieves GF F1 of 0.613 at cap=0.6, closely approaching the no-compression baseline of 0.637, while ParallelComp yields only 0.297.
SnapKV performs competitively on file-level localization and achieves higher GF F1 than CodeComp on Qwen-Coder at cap=0.6 (0.700 vs. 0.613); however, CodeComp consistently produces patches with lower edit distance (0.683 vs. 0.718), indicating better generation fidelity.
While SnapKV achieves higher GF F1 on Qwen-Coder at cap=0.6, CodeComp produces patches with lower edit distance (0.683 vs. 0.718), suggesting that the two methods optimize for different aspects of the task. The absolute resolve rate remains modest across all methods; patch edit distance under CodeComp closely matches the uncompressed baseline (0.743 vs. 0.740 on DS-Coder at cap=0.4), indicating that compression quality is not the primary limiting factor.

\begin{table}[t]
\centering
\small
\setlength{\tabcolsep}{5pt}
\begin{tabular}{l@{\hspace{5pt}}l@{\hspace{6pt}}c c c c c c c}
\toprule
Model & Method & Cap. & GF Prec & GF Rec & GF F1 & GF Jac & Patch Valid & Edit Dist \\
\midrule
DS-Coder & No Comp. & 1.0 & 0.338 & 0.338 & 0.338 & 0.338 & 0.875 & 0.740 \\
DS-Coder & SnapKV & 0.4 & 0.200 & 0.200 & 0.200 & 0.200 & 0.900 & 0.758 \\
DS-Coder & ParallelComp & 0.4 & 0.019 & 0.025 & 0.021 & 0.019 & 0.350 & 0.824 \\
\rowcolor{blue!8} DS-Coder & CodeComp & 0.4 & 0.250 & 0.250 & 0.250 & 0.250 & 1.000 & \textbf{0.743} \\
DS-Coder & SnapKV & 0.6 & 0.350 & 0.350 & 0.350 & 0.350 & 0.800 & 0.770 \\
DS-Coder & ParallelComp & 0.6 & 0.038 & 0.038 & 0.038 & 0.038 & 0.225 & 0.869 \\
\rowcolor{blue!8} DS-Coder & CodeComp & 0.6 & 0.250 & 0.250 & 0.250 & 0.250 & 0.888 & \textbf{0.766} \\
\midrule
Qwen-Coder & No Comp. & 1.0 & 0.625 & 0.650 & 0.637 & 0.625 & 1.000 & 0.666 \\
Qwen-Coder & SnapKV & 0.4 & 0.600 & 0.600 & 0.600 & 0.600 & 0.950 & 0.709 \\
Qwen-Coder & ParallelComp & 0.4 & 0.369 & 0.375 & 0.372 & 0.369 & 0.663 & 0.800 \\
\rowcolor{blue!8} Qwen-Coder & CodeComp & 0.4 & 0.544 & 0.550 & 0.547 & 0.544 & 0.988 & \textbf{0.686} \\
Qwen-Coder & SnapKV & 0.6 & 0.700 & 0.700 & 0.700 & 0.700 & 1.000 & 0.718 \\
Qwen-Coder & ParallelComp & 0.6 & 0.294 & 0.300 & 0.297 & 0.294 & 0.663 & 0.809 \\
\rowcolor{blue!8} Qwen-Coder & CodeComp & 0.6 & 0.613 & 0.613 & 0.613 & 0.613 & 1.000 & \textbf{0.683} \\
\bottomrule
\end{tabular}
\caption{Results on SWE-bench Lite. GF denotes ground-truth 
file-level evaluation. Cap. denotes the KV cache capacity 
ratio.}
\label{tab:swebench_lite}
\end{table}
\begin{table}[t]
\centering
\small
\setlength{\tabcolsep}{6pt}
\begin{tabular}{l@{\hspace{5.6pt}}l@{\hspace{6pt}}c c c c c c}
\toprule
Model & Method & Cap. & InfBench & DebugBench & LongCodeQA & Str. Score & Avg. \\
\midrule
Llama3-8B & No Comp.     & 1.0 & 0.20 & 0.74 & 0.59 & 1.00 & 0.51 \\
Llama3-8B & SnapKV       & 0.4 & 0.10 & 0.41 & 0.48 & 0.36 & 0.33 \\
Llama3-8B & ParallelComp & 0.4 & 0.10 & 0.03 & 0.38 & 0.35 & 0.17 \\
\rowcolor{blue!8}
Llama3-8B & CodeComp     & 0.4 & \textbf{0.14} & \textbf{0.43} & \textbf{0.52} & \textbf{0.54} & \textbf{0.36} \\
Llama3-8B & SnapKV       & 0.6 & 0.12 & 0.44 & \textbf{0.56} & 0.57 & 0.37 \\
Llama3-8B & ParallelComp & 0.6 & 0.11 & 0.17 & 0.31 & \textbf{0.58} & 0.20 \\
\rowcolor{blue!8}
Llama3-8B & CodeComp     & 0.6 & \textbf{0.18} & \textbf{0.49} & 0.54 & \textbf{0.75} & \textbf{0.40} \\
\midrule
Qwen3-8B  & No Comp.     & 1.0 & 0.21 & 0.87 & 0.67 & 1.00 & 0.58 \\
Qwen3-8B  & SnapKV       & 0.4 & \textbf{0.17} & 0.71 & \textbf{0.69} & 0.36 & \textbf{0.52} \\
Qwen3-8B  & ParallelComp & 0.4 & 0.10 & 0.25 & 0.13 & 0.33 & 0.16 \\
\rowcolor{blue!8}
Qwen3-8B  & CodeComp     & 0.4 & 0.16 & \textbf{0.72} & 0.64 & \textbf{0.56} & 0.51 \\
Qwen3-8B  & SnapKV       & 0.6 & 0.17 & 0.66 & \textbf{0.68} & 0.57 & 0.50 \\
Qwen3-8B  & ParallelComp & 0.6 & 0.05 & 0.48 & 0.17 & 0.52 & 0.23 \\
\rowcolor{blue!8}
Qwen3-8B  & CodeComp     & 0.6 & \textbf{0.21} & \textbf{0.79} & 0.58 & \textbf{0.77} & \textbf{0.53} \\
\bottomrule
\end{tabular}
\caption{Bug localization accuracy on InfiniteBench-CodeDebug (InfBench), DebugBench, 
and LongCodeQA under different retention budgets. Str.\ Score denotes the average 
fraction of structurally critical tokens retained across layers and chunks, 
computed via CPG-derived span annotations. Best results under each compression 
ratio are \textbf{bolded}.}
\label{tab:bug_localization}
\end{table}

\subsection{Bug Localization}
Table~\ref{tab:bug_localization} reports results on InfiniteBench-CodeDebug~\citep{zhang-etal-2024-bench}, DebugBench~\citep{tian2024debugbench}, and LongCodeQA~\citep{rando2025longcodebenchevaluatingcodingllms}. CodeComp consistently outperforms both ParallelComp and SnapKV across all models, benchmarks, and retention budgets, while also retaining a substantially higher fraction of structurally critical tokens (Str.\ Score). The gap over ParallelComp is most pronounced on DebugBench: at cap=0.4, ParallelComp collapses to 0.03 on Llama3-8B and 0.25 on Qwen3-8B, while CodeComp achieves 0.43 and 0.72 respectively (\textbf{14$\times$} and \textbf{2.9$\times$} improvements), with a correspondingly higher Str.\ Score (0.54 vs.\ 0.35). On Qwen3-8B at cap=0.6, CodeComp achieves an average of 0.53, recovering 91\% of the no-compression baseline (0.58), while ParallelComp reaches only 0.23. CodeComp also outperforms SnapKV on most settings. Notably, despite SnapKV and CodeComp retaining similar numbers of tokens, CodeComp achieves higher Str.\ Score (e.g., 0.77 vs.\ 0.57 on Qwen3-8B at cap=0.6), confirming that structural span protection targets more semantically relevant regions than observation-window eviction. 

Across all settings, CodeComp advances the Pareto frontier of accuracy versus memory efficiency, consistently achieving higher accuracy than attention-only baselines at the same retention budget and approaching full-context performance at significantly reduced 
memory cost.

\subsection{Ablation}
As shown in  Table~\ref{tab:ablation_span_capacity}, our ablation reveals a clear functional separation between the two components. Span-level preservation is the dominant contributor in our ablation. Across all compression budgets, the span-only variant consistently and substantially outperforms the capacity-only variant, indicating that preserving structurally meaningful local code regions is much more important than merely redistributing budget across chunks. This also suggests that, for debugging-oriented code tasks, the main bottleneck lies in retaining the right intra-chunk evidence rather than only identifying which chunk should receive more tokens.

Chunk-level budget reallocation provides consistent but modest additional benefit over span-only: the full model (Capacity-Span) matches or slightly exceeds span-only across all settings, with the best results achieved under stronger allocation strengths (e.g., 0.25/1.75 and 0.1/1.9 at cap=0.6). This suggests that budget allocation and span protection are complementary, with span-level evidence preservation serving as the primary mechanism and budget reallocation providing a lightweight refinement.

\begin{figure}[h]
    \centering
    \begin{subfigure}[t]{0.48\columnwidth}
        \includegraphics[width=\linewidth]{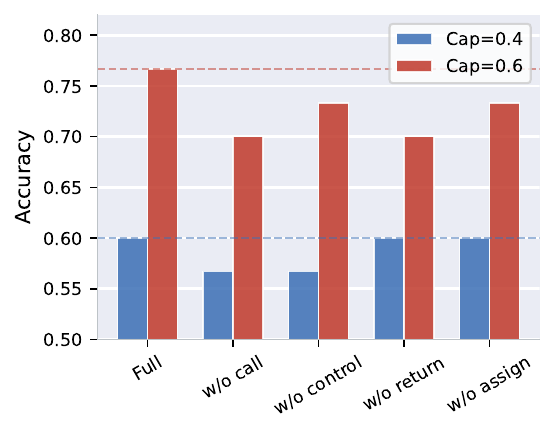}
        \caption{Feature ablation of span scoring.}
        \label{fig:ablation_feature}
    \end{subfigure}
    \hfill
    \begin{subfigure}[t]{0.48\columnwidth}
        \includegraphics[width=\linewidth]{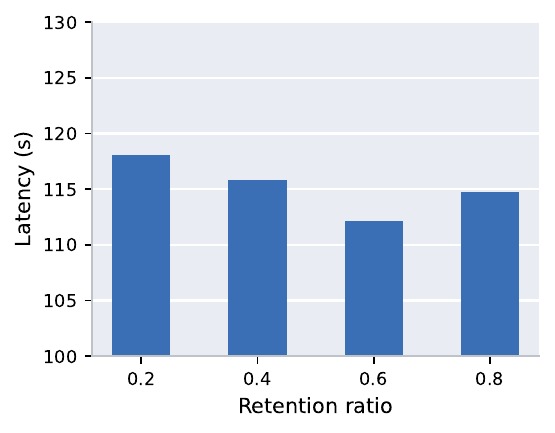}
        \caption{End-to-end latency on SWE-bench Lite.}
        \label{fig:throughput}
    \end{subfigure}
    \caption{Feature-level ablation and throughput analysis.}
    \label{fig:ablation_throughput}
\end{figure}

\begin{table}[t]
\centering
\small
\setlength{\tabcolsep}{10pt}
\begin{tabular}{clcccccc}
\toprule
\textbf{Cap} & \textbf{Method} & \textbf{0.9/1.1} & \textbf{0.75/1.25} & \textbf{0.5/1.5} & \textbf{0.25/1.75} & \textbf{0.1/1.9} & \textbf{Best} \\
\midrule

\multirow{3}{*}{0.2} 
& Capacity-only & 0.433 & 0.400 & 0.417 & 0.400 & 0.450 & 0.450 \\
& Span-only     &  --   &  --   &  --   &  --   &  --   & 0.617 \\
\rowcolor{blue!8} \cellcolor{white}
& Capacity-Span          & 0.617 & 0.650 & 0.633 & 0.633 & 0.600 & \textbf{0.650} \\
\midrule

\multirow{3}{*}{0.4} 
& Capacity-only & 0.283 & 0.250 & 0.233 & 0.267 & 0.250 & 0.283 \\
& Span-only     &  --   &  --   &  --   &  --   &  --   & 0.633 \\
\rowcolor{blue!8} \cellcolor{white}
& Capacity-Span          & 0.650 & 0.633 & 0.667 & 0.633 & 0.650 & \textbf{0.667} \\
\midrule

\multirow{3}{*}{0.6} 
& Capacity-only & 0.417 & 0.450 & 0.383 & 0.400 & 0.400 & 0.450 \\
& Span-only     &  --   &  --   &  --   &  --   &  --   & 0.783 \\
\rowcolor{blue!8} \cellcolor{white}
& Capacity-Span          & 0.783 & 0.783 & 0.783 & 0.800 & 0.800 & \textbf{0.800} \\

\bottomrule
\end{tabular}
\caption{Ablation results on DebugBench with 60 samples under different global compression ratios and budget-allocation strengths. Span-level preservation provides the main gains, while chunk-level budget allocation alone is weak. Capacity-Span benefits more under stronger allocation settings.}
\label{tab:ablation_span_capacity}
\end{table}
Figure~\ref{fig:ablation_feature} provides a feature-level ablation of the span scoring function. Removing \textit{call} or \textit{control} leads to the most consistent degradation, while removing \textit{return} causes a noticeable drop at cap=0.6. Removing \textit{assign} has a smaller effect. These results indicate that span-level protection derives its effectiveness primarily from preserving call and control-flow anchors.

% \subsection{Throughput}
% \input{table/throughput}
% Table~\ref{tab:throughput_debugbench} reports end-to-end latency and throughput on DebugBench. CodeComp remains broadly competitive with ParallelComp in efficiency under both compression budgets. 
% At the more aggressive setting (\texttt{Cap.=0.4}), CodeComp incurs a modest latency overhead, which is expected because stronger compression requires more fine-grained structural planning and span protection within a tighter budget. At \texttt{Cap.=0.6}, the efficiency gap largely disappears, suggesting that the additional structural processing cost is limited and does not substantially affect end-to-end throughput under a less restrictive cache budget.

\subsection{Throughput}
% \begin{wrapfigure}{r}{0.4\columnwidth}
%     \vspace{-30pt}
%     \centering
% \includegraphics[width=\linewidth]{figure/throughput_latency.pdf}
%     \caption{End-to-end latency of CodeComp on SWE-bench 
%     Lite across retention ratios.}
%     \label{fig:throughput}
%     \vspace{-20pt}
% \end{wrapfigure}
Figure~\ref{fig:throughput} reports end-to-end latency of CodeComp on SWE-bench Lite with Qwen2.5-Coder-32B-Instruct across different retention ratios. Latency remains stable across all settings (112--118s) , indicating that the structural analysis overhead introduced by Joern and the span protection mechanism does not substantially affect end-to-end inference efficiency. Unlike existing compression methods whose official implementations are based on the Transformers library, CodeComp is implemented natively on SGLang, enabling seamless integration into high-throughput agentic coding pipelines.

% \vspace{-5pt}
% \section{Limitation}
% The span importance weights in CodeComp are currently set based on structural type and kept fixed across tasks. While this simple weighting scheme proves effective in our experiments, future work could explore data-driven approaches to further adapt these weights to different tasks or codebases. We also observe that chunk-level budget allocation provides consistent but modest benefits in the current design; a more principled allocation strategy could potentially amplify its contribution.
\vspace{-8pt}
\section{Conclusion}

We presented CodeComp, a structure-aware KV cache compression framework that incorporates static program analysis into LLM inference for repository-level code tasks. Implemented on top of SGLang~\citep{zheng2024sglangefficientexecutionstructured}, CodeComp is designed for efficient, large-scale deployment and is compatible with agentic inference workflows that require high-throughput generation over long code contexts. By extracting structural priors from Code Property Graphs via Joern, CodeComp guides two complementary compression decisions: span-level protection, which explicitly preserves semantically critical local code regions such as call expressions, control-flow predicates, and return statements, and structure-aware budget allocation, which distributes compression capacity according to chunk-level structural importance. Empirical results on bug localization and code generation benchmarks demonstrate that CodeComp consistently and substantially outperforms attention-only compression baselines under the same memory budget, while matching the patch generation quality of uncompressed full-context inference. Ablation studies confirm that span-level protection is the primary source of gains, with call and control-flow anchors being the most critical structural signals. We also introduce a structure score metric that quantifies the fraction of structurally critical tokens retained under compression; this metric correlates strongly with task accuracy across methods and settings, suggesting it serves as a reliable proxy for compression quality in code-centric tasks. We hope CodeComp serves as a foundation for future work on structure-aware inference optimization for agentic software engineering.
\newpage
\bibliography{colm2026_conference}
\bibliographystyle{colm2026_conference}

\appendix
\section{Appendix}
\subsection{Experimental Setup}

\paragraph{Models and benchmarks.}
We evaluate two 8B-scale code language models, Llama-3-8B-Instruct~\citep{llama3modelcard} and Qwen3-8B~\citep{qwen3technicalreport}, across five benchmarks spanning two task types. For bug localization, we use InfiniteBench-CodeDebug~\citep{zhang-etal-2024-bench}, DebugBench~\citep{tian2024debugbench}, and LongCodeQA~\citep{rando2025longcodebenchevaluatingcodingllms}. For code generation, we use SWE-bench Lite~\citep{jimenez2024swebenchlanguagemodelsresolve} and the LCA library-based code generation benchmark~\citep{bogomolov2024long}. All experiments are conducted using SGLang~\citep{zheng2024sglangefficientexecutionstructured} (v0.4.9.post3), a high-throughput inference engine designed for agentic LLM workflows, enabling efficient large-scale evaluation.

\paragraph{Compression settings.}
We evaluate three conditions: \textit{No compression} (full KV cache, capacity = 1.0), \textit{ParallelComp} (attention-based chunk compression without structural guidance, capacity $\in \{0.4, 0.6\}$), and \textit{CodeComp} (using the same capacity settings). The context is partitioned into chunks of up to 4096 tokens, with a target of 512 tokens per file-level chunk and a minimum of 128 tokens per chunk. Attention scores are aggregated using average pooling over a sliding window of size 5, with a context window of 128 tokens.

\paragraph{Structural prior configuration.}
CodeComp enables span-level protection with a span budget ratio of 0.5 relative to the chunk token budget, a minimum span size of 16 tokens, and a merge gap of 1 line. Span importance scores are computed as a weighted combination of structural features: function calls (0.20), control-flow constructs (0.18), query alignment (0.18), return statements (0.14), assignments (0.14), def-use edges (0.10), and attention signal (0.06). For chunk-level capacity reallocation, structural scores are mapped to capacity multipliers in the range $[0.5, 1.5]$.

\paragraph{Dataset-specific settings.}
SWE-bench Lite and LCA CodeGen use an additional BM25-based file retrieval stage (top-20 files) prior to chunk selection. InfiniteBench-CodeDebug and LongCodeQA use top-6 chunk selection with a maximum of 64 new tokens. DebugBench uses top-4 chunk selection with a maximum of 512 new tokens to accommodate full bug-fix outputs. All other parameters are held constant across datasets and models, so the primary factors of variation are the model, the method, and the compression ratio.

\subsection{LCA CodeGen Benchmark}

\begin{table*}[t]
\centering
\small
\setlength{\tabcolsep}{8.3pt}
\begin{tabular}{l@{\hspace{8pt}}l@{\hspace{6pt}}c c c c c c}
\toprule
Model & Method & Cap. & API Prec & API Rec & API F1 & API Jac & Edit Dist \\
\midrule
DS-Coder & No Comp. & 1.0 & 0.151 & 0.242 & 0.186 & 0.118 & 0.601 \\
DS-Coder & ParallelComp & 0.4 & 0.109 & 0.162 & 0.131 & 0.081 & 0.712 \\
DS-Coder & SnapKV & 0.4 & 0.048 & 0.365 & 0.085 & 0.044 & 0.716 \\
\rowcolor{blue!8} DS-Coder & CodeComp & 0.4 & \textbf{0.176} & \textbf{0.239} & \textbf{0.202} & \textbf{0.124} & \textbf{0.628} \\
DS-Coder & ParallelComp & 0.6 & 0.132 & 0.154 & 0.142 & 0.086 & 0.705 \\
DS-Coder & SnapKV & 0.6 & 0.050 & 0.408 & 0.088 & 0.047 & 0.722 \\
\rowcolor{blue!8} DS-Coder & CodeComp & 0.6 & \textbf{0.193} & \textbf{0.298} & \textbf{0.234} & \textbf{0.147} & \textbf{0.600} \\
\midrule
Qwen-Coder & No Comp. & 1.0 & 0.158 & 0.274 & 0.201 & 0.125 & 0.581 \\
Qwen-Coder & ParallelComp & 0.4 & 0.041 & 0.386 & 0.074 & 0.038 & 0.742 \\
Qwen-Coder & SnapKV & 0.4 & 0.052 & 0.434 & 0.094 & 0.050 & 0.737 \\
\rowcolor{blue!8} Qwen-Coder & CodeComp & 0.4 & \textbf{0.054} & \textbf{0.554} & \textbf{0.100} & \textbf{0.052} & \textbf{0.665} \\
Qwen-Coder & ParallelComp & 0.6 & 0.034 & 0.315 & 0.068 & 0.032 & 0.748 \\
Qwen-Coder & SnapKV & 0.6 & 0.052 & 0.502 & 0.095 & 0.050 & 0.733 \\
\rowcolor{blue!8} Qwen-Coder & CodeComp & 0.6 & \textbf{0.054} & \textbf{0.567} & \textbf{0.099} & \textbf{0.052} & \textbf{0.670} \\
\bottomrule
\end{tabular}
\caption{Results on LCA CodeGen. API denotes API-level 
correctness. Cap. denotes the KV cache retention ratio. }
\label{tab:lca_codegen}
\end{table*}

Table~\ref{tab:lca_codegen} reports additional results on LCA CodeGen. CodeComp consistently outperforms both ParallelComp and SnapKV across both models and retention budgets, achieving higher API precision, F1, and Jaccard similarity while also reducing edit distance. On DS-Coder at cap=0.6, CodeComp achieves API F1 of 0.234 versus 0.142 for ParallelComp and 0.088 for SnapKV, and reduces edit distance to 0.600, matching the uncompressed baseline of 0.601. Notably, SnapKV achieves relatively high API recall but with very low precision, suggesting it generates overly broad outputs that superficially match API references without structural correctness. These results further confirm that structure-aware retention is beneficial for code generation tasks requiring precise preservation of semantically relevant repository context.

% \subsection{Position Encoding}
% \label{sec:position_encoding}

% \textbf{PPL-based chunk scoring.}
% During the chunk selection stage, each candidate chunk $C_i$ is scored independently. The input to the model is formatted as $[\text{prefix}; C_i; \text{query}]$, where position indices are assigned contiguously starting from 0. This allows parallel scoring of all chunks without positional interference between them.

% \textbf{Compression and decode.}
% After top-$k$ chunks are selected and compressed, the final input is assembled as:
% \begin{equation}
% [\text{prefix};\; \hat{C}_1;\; \hat{C}_2;\; \dots;\; 
% \hat{C}_k;\; \text{query}]
% \end{equation}
% where $\hat{C}_i$ denotes the compressed version of chunk $C_i$. Each chunk retains its original position indices from the scoring stage. Since chunks are scored independently, their position indices may overlap. To avoid positional conflict with the query, we assign the query a starting 
% position of:
% \begin{equation}
% p_{\text{query}} = \max_{i} \operatorname{len}(C_i) + 
% \operatorname{len}(\text{prefix})
% \end{equation}
% where $\operatorname{len}(C_i)$ is the token length of chunk 
% $C_i$ before compression. This ensures that the query always receives the highest position indices in the sequence, consistent with its role as the decoding target, regardless of the actual lengths of the compressed chunks.

\end{document}